\PassOptionsToPackage{most}{tcolorbox}

\documentclass[10pt,letterpaper]{mystyle}

\usepackage{float}
\usepackage{placeins}
\usepackage{url}
\usepackage{natbib}
\usepackage{fvextra}
\usepackage{graphicx}
\usepackage{booktabs}
\usepackage{multirow}
\usepackage{subcaption}
\usepackage{wrapfig}
\usepackage{listings}
\usepackage{siunitx}
\usepackage{amsthm}
\usepackage{amsfonts}
\usepackage{bbm}
\usepackage{fontawesome5} 
\usepackage{listings}
\usepackage[textsize=tiny]{todonotes}
\usepackage{tikz}
\usepackage{multicol}
\usetikzlibrary{arrows.meta, shapes, shapes.geometric, positioning} 

\usepackage{tcolorbox} 

\definecolor{usercolor}{RGB}{230, 242, 255} 
\definecolor{usertitle}{RGB}{180, 210, 255} 
\definecolor{agentcolor}{RGB}{245, 245, 245} 
\definecolor{agenttitle}{RGB}{220, 220, 220} 
\definecolor{statecolor}{RGB}{255, 253, 220} 
\definecolor{stateborder}{RGB}{200, 180, 50} 

\usepackage[most]{tcolorbox}
\usepackage{fontawesome5} 
\usepackage{xcolor}

\usepackage{amsmath,amsfonts,bm}

\def\eqref#1{equation~\ref{#1}}

\def\1{\bm{1}}

\DeclareMathAlphabet{\mathsfit}{\encodingdefault}{\sfdefault}{m}{sl}
\SetMathAlphabet{\mathsfit}{bold}{\encodingdefault}{\sfdefault}{bx}{n}

\theoremstyle{plain}

\newtheoremstyle{sig}
  {} {} {\itshape} {} {\scshape} {.} {.5em} {#1 #2\thmnote{\quad(#3)}}
\theoremstyle{sig}

\definecolor{dark2green}{rgb}{0.1, 0.65, 0.3}
\definecolor{dark2orange}{rgb}{0.9, 0.4, 0.}
\definecolor{dark2purple}{rgb}{0.4, 0.4, 0.8}
\definecolor{c1}{HTML}{9B5353}
\definecolor{c3}{HTML}{9B5353}

\usepackage{hyperref}
\hypersetup{
    colorlinks=true,
    linkcolor=c3,
    urlcolor=c3,
    citecolor=black,
}
\usepackage[capitalize,noabbrev]{cleveref}

\title{Consistent but Miscalibrated: Evaluating LLM Limitations for Risk Communication in Natural Language}
\runningtitle{Consistent but Miscalibrated: Evaluating LLM Limitations for Risk Communication in Natural Language}

\author[1,2,3]{Diego Cerda-Mardini}
\author[2,3,4,5]{Sarath Chandar}
\author[1,3]{Sreenath Madathil}

\affil[1]{Faculty of Dental Medicine and Oral Health Sciences, McGill University, Montréal, QC, Canada}
\affil[2]{Chandar Research Lab, Polytechnique Montréal, Montréal, H3T 1J4, Canada}
\affil[3]{Mila – Québec Artificial Intelligence Institute, Montréal, H2S 3H1, Canada}
\affil[4]{Département de Génie Informatique et Génie Logiciel (GIGL), Polytechnique Montréal, Montréal, H3T 1J4, Canada}
\affil[5]{Canada CIFAR AI Chair}

\correspondingauthor{Sreenath Madathil \href{mailto:sreenath.madathil@mcgill.ca}{sreenath.madathil@mcgill.ca}}

\begin{document}

\begin{abstract}

LLMs are increasingly deployed as post-hoc explainers of AI-generated outputs, yet it remains unclear whether they can reliably communicate probabilistic information in natural language. For this role to be viable, models must produce identical verbal descriptions for identical inputs, and select descriptions that accurately reflect the magnitude of the underlying numerical quantities. We evaluate whether nine LLMs meet these requirements within a two-stage prediction pipeline, in which an upstream model has produced probabilistic outputs characterized by their likelihood and uncertainty, and LLMs are tasked with selecting an appropriate verbal descriptor for each. We simulate predictions from an upstream model by taking samples from a Beta distribution parameterized by its mode and prior sample size. We then prompt LLMs to explain these predictions under six domain contexts and with ten temperature settings, and repeating each experiment ten times. We find that LLMs are generally consistent but miscalibrated, with substantially weaker performance on uncertainty than on likelihood tasks. Providing models with precomputed summary statistics (mode and prior sample size) reduced sensitivity to contextual framing but did not resolve the underlying miscalibration, suggesting that the bottleneck resides in the verbalization step itself. These findings indicate that current LLMs do not yet constitute reliable zero-shot standalone risk communication tools for probabilistic predictions.

\includegraphics[height=1em]{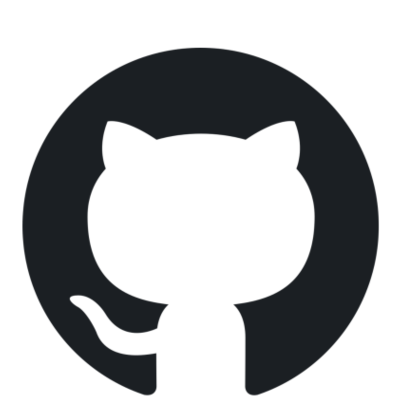} \textbf{Code Repository}: \href{https://github.com/CTruAI/risk_communication_llm}{\textbf{github/CTruAI/risk\_communication\_llm}}

\end{abstract}
\maketitle

\section{Introduction}\vspace{-0.25cm}
\label{sec:Introduction}

\begin{multicols}{2}

Large Language Models (LLMs) are increasingly used in high-stakes domains, such as healthcare, finance and law, where the cost of incorrect predictions can be severe \citep{Maity2025LLMHealthcare, Dehghani2025LLMLegal, Nie2024LLMFinance}. In such settings, it is not enough for a model to be accurate; users also need to know how confident the model is in its predictions \citep{Band2024}. Calibrated uncertainty scores address this need by indicating when to trust a model's prediction \citep{Minderer2021, Madsen2024}. Yet even when such scores are available, correctly interpreting them poses a separate challenge for end users. Concepts like ``statistical uncertainty'' are frequently misinterpreted, even by skilled professionals like doctors \citep{Gigerenzer2007}. Predictive distributions can be characterized by two properties: the likelihood of an event occurring and the uncertainty in its prediction \citep{Tyralis2024}. How laypeople intuitively understand probability often diverges from its formal statistical definition \citep{Hashim2024}, adding to the misunderstanding of risk. As reliance on AI for decision-making grows, there is need to convey predictive concepts like likelihood and uncertainty to audiences with varying statistical literacy. LLMs are a promising vehicle for this, as they may be able to translate numerical estimates into Natural Language Explanations (NLEs) that are more interpretable to non-expert users \citep{kayser_emde_camburu_parsons_papiez_lukasiewicz_2022, Kayser2021eViLAD, Stern2024NaturalLE}. This raises two questions: whether LLMs have a coherent internal mapping of concepts like likelihood and uncertainty, and whether they can express these concepts in language that is consistent and calibrated to the underlying probabilistic information. To address this, we evaluated nine LLMs on their ability to produce NLEs from simulated predictive distributions, assessing their suitability for risk communication through consistency and calibration metrics.

\end{multicols}
\newpage
\section{Definitions}\vspace{-0.25cm}
\label{sec:Definitions}

\begin{multicols}{2}

\textbf{Likelihood} refers to the probability of an event occurring. In this context, it is taken as the mode of the distribution assumed to be from a probabilistic prediction model \citep{Murphy2022}.

\noindent \textbf{Predictive uncertainty} refers to the degree of uncertainty around a prediction, typically derived from the variance of predicted probabilities in uncertainty-aware models \citep{KendallGal2017}.

\noindent \textbf{Natural Language Explanations (NLEs)} are verbal descriptions of numerical inputs, used in explainable AI to make model outputs interpretable. Here, NLEs take the form of verbal descriptors of likelihood and uncertainty, such as "very likely" or "highly uncertain"\citep{kayser_emde_camburu_parsons_papiez_lukasiewicz_2022}.

\noindent \textbf{Calibration} is the degree to which a model's chosen descriptor accurately reflects the magnitude of a known numerical quantity. A well-calibrated model selects stronger descriptors for higher values and weaker descriptors for lower ones. For example, preferring "very likely" over "unlikely" for a high-probability prediction. This concept is adapted from the notion of ``epistemic calibration'', defined as the alignment between a model's internal certainty and its linguistic assertiveness \citep{Ghafouri2025}.

\noindent \textbf{Consistency} is the degree to which a model produces identical descriptors for identical inputs across repeated queries \citep{Kolena2024}.

\noindent \textbf{Risk communication} is the practice of presenting probabilistic information in a transparent and interpretable manner to support informed decision-making \citep{Spiegelhalter2011}. In this study, it is operationalized as the accurate verbal description of the likelihood and uncertainty of a predictive distribution, requiring both calibration and consistency.

\end{multicols}
\section{Related works}\vspace{-0.25cm}
\label{sec:Related works}

\begin{multicols}{2}

\subsection{The challenge of risk communication}\vspace{-0.25cm}
Numerical probabilities are precise but frequently misinterpreted by low-numeracy populations, while verbal descriptions offer better intuitive access but vary widely in how people interpret the same words \citep{Spiegelhalter2011}. This mismatch extends beyond lay audiences: clinicians often lack formal training in statistical reasoning and struggle to interpret risk estimates \citep{Hashim2024, Gigerenzer2007}. Substantial vocabulary mismatches have been documented between expert panels and the public when interpreting probabilities \citep{Jackson2025}. The consequences of poor risk communication in high-stakes domains can be far-reaching. For example, when the International Agency for Research on Cancer (IARC) classified processed meat as a Group 1 carcinogen in 2015, the announcement provoked widespread alarm largely because the distinction between hazard classification (whether an agent \textit{can} cause cancer) and risk magnitude (how \textit{much} it increases cancer probability) was not made clear \citep{bouvard_loomis_guyton_grosse_ghissassi_benbrahim-tallaa_guha_mattock_straif_2015, gallus_bosetti_2016}. Such episodes illustrate that even institutional experts can fail to translate probabilistic concepts into language the public interprets correctly. As AI-driven predictions become more prevalent, there is growing need to evaluate the abilities and limitations of LLMs for probabilistic risk communication.
\vspace{-0.66cm}

\subsection{Defining calibration for probabilistic language}\vspace{-0.25cm}
Several related concepts have been proposed to characterize how well a model's language reflects its underlying probabilistic state, or its alignment to an external ground truth. \citet{Minderer2021} defined ``calibrated uncertainty'' as the accuracy with which scores provided by a model reflect its predictive uncertainty. Adapting this to natural language, \citet{Ghafouri2025} introduced ``epistemic calibration'', defined as the alignment between a model's internal confidence and the way it expresses that confidence linguistically. \citet{Eikema+2025} proposed a closely related concept called ``faithful uncertainty'', which is when a model's communicated decisiveness matches its inner state of predictive uncertainty. These concepts have primarily been evaluated on short form LLM outputs. For long form generation, \citet{Band2024} proposed ``linguistic calibration'', which equates calibration to usefulness by determining whether a confidence statement allows users to make calibrated probabilistic predictions.

\subsection{Evaluation approaches for probabilistic verbalization}\vspace{-0.25cm}
Different metrics have been proposed for measuring how well LLMs communicate predictive properties in natural language. \citet{Jackson2025} measured abstention rates, or how often LLMs fail to provide verbalized interpretations of numerical inputs, and provided mappings between NLE descriptors and corresponding numerical probabilities. \citet{Eikema+2025} defined probability hedges, which are ordered sequences of words ranging from less to more probable, and evaluated the capacity of LLMs to assign associated probabilities to these hedges. Complementary frameworks have also been proposed for quantifying explanation level uncertainty, either by prompting the model to express its own confidence (verbalized uncertainty) or by measuring variation in outputs under repeated sampling and model perturbations (probing uncertainty) \citep{harsha_agarwal_lakkaraju_2023}, as well as for evaluating the calibration of confidence statements in long form outputs through the lens of downstream usefulness \citep{Band2024}. While these approaches capture important dimensions of probabilistic verbalization, they have generally been applied in isolation, evaluating either abstention behavior or probability to hedge mappings, rather than jointly assessing calibration and consistency under controlled experimental conditions. More importantly, none of these studies have looked at probabilistic verbalization in the context of LLM as post-hoc explainers.

\subsection{Empirical findings on LLMs and probabilistic language}\vspace{-0.25cm}
Several studies have examined LLM behavior with probabilistic information, revealing systematic limitations. \citet{Ulmer+2025} found that models tend to prioritize helpfulness and decisiveness at the expense of expressing uncertainty, resulting in limited use of hedging language, which are words or phrases that soften a statement, such as ``somewhat'' or ``possibly''. \citet{Eikema+2025} found considerable inconsistencies in hedging vocabulary and proposed Faithful Uncertainty Tuning (FUT), a finetuning framework shown to improve hedging calibration. \citet{harsha_agarwal_lakkaraju_2023} investigated verbalized confidence in Chain-of-Thought (CoT) explanations, finding that LLMs consistently report near-maximum confidence regardless of explanation quality, averaging 94\% across tasks, reinforcing the concern that LLM language about uncertainty does not reliably reflect underlying information. \citet{Jackson2025} found that LLMs are less likely to provide probability explanations when scenarios are more severe or when users express anxious vocabulary, suggesting that models respond to emotional tone rather than underlying data.

Together, these findings point to critical vulnerabilities in the capacity of LLMs to produce calibrated descriptions of probabilistic information. However, existing evaluations have generally examined individual aspects, such as hedging behavior, verbalized confidence, or abstention, in isolation, often on few models or under limited experimental variation. A controlled evaluation that jointly measures calibration and consistency across multiple architectures, domain contexts, and generation settings remains lacking, and is the focus of the present study.

\end{multicols}
\section{Experimental setup}\vspace{-0.25cm}
\label{sec:Experimental setup}

\begin{multicols}{2}

\subsection{Problem statement}
\begin{figure*}[t]
  \begin{center}
  \includegraphics[width=1\linewidth]{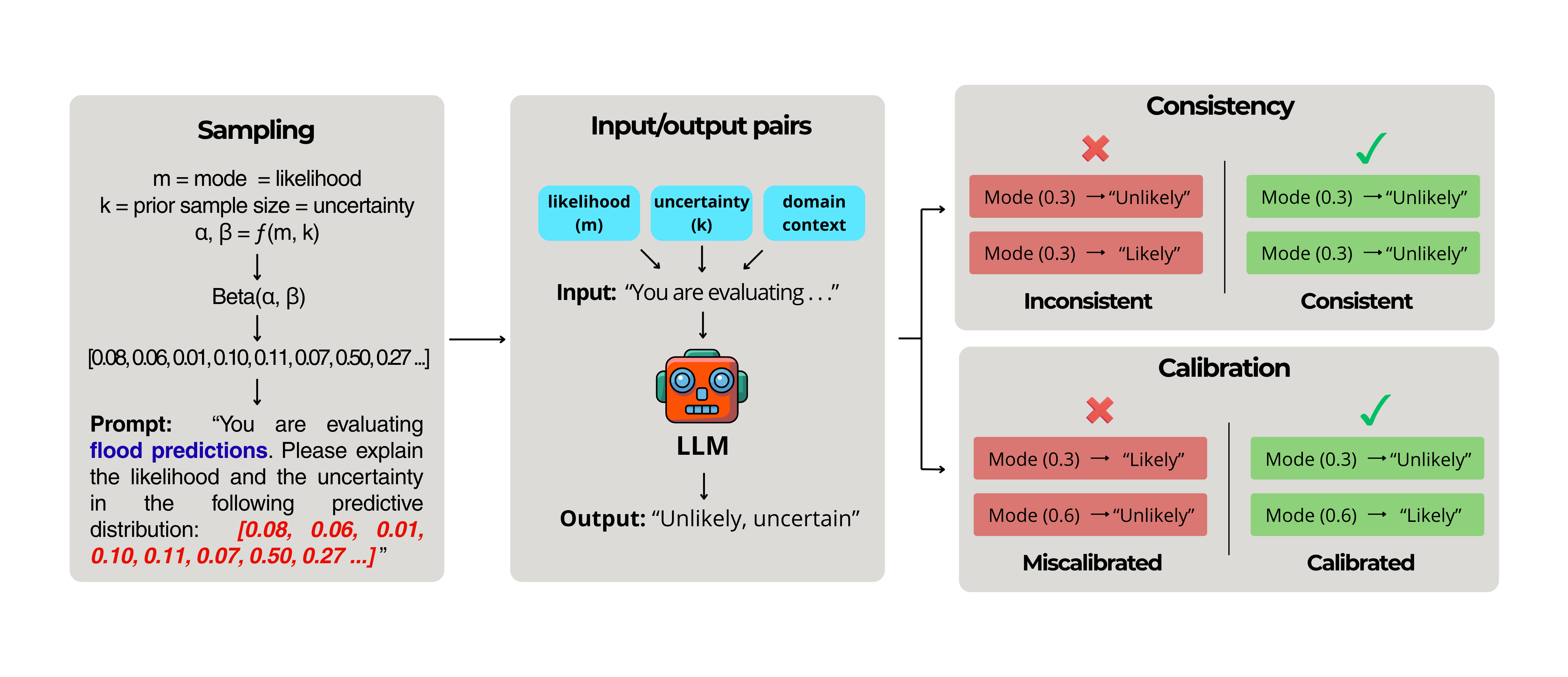}
  \end{center}
  \caption{Samples were drawn from a Beta distribution with parameters m (mode) and k (prior sample size). These samples were then inserted into a prompt containing a domain context, as well as task instructions. LLMs were prompted to choose the best NLE descriptor for the likelihood and uncertainty of these simulated scenarios, as can be seen in the input-output pairs panel. LLMs were then evaluated for consistency and calibration scores across experiments.}
  \label{fig:experiments}
\end{figure*}

\vspace{-0.25cm}
\label{sec:Problem statement}
 We formalize risk communication as a descriptor selection task. We assume predictions are generated by an upstream AI or statistical model, whose output takes the form of a set of probability samples (as is common in uncertainty-aware approaches such as Monte Carlo Dropout \citep{gal2016}), from which likelihood and uncertainty estimates are derived. Given a predictive distribution characterized by these quantities, can an LLM reliably select verbal descriptors that accurately reflect the underlying numerical values? We evaluate this along two dimensions: whether models produce the same descriptor for identical inputs across repetitions (consistency), and whether their descriptor choices respect the numerical ordering of the underlying distributions (calibration). Both properties are necessary for an LLM to serve as a reliable risk communication tool, as inconsistency undermines trustworthiness while miscalibration undermines accuracy. Figure~\ref{fig:experiments} summarizes our experiment setup.

\subsection{Task formulation and experimental design}\vspace{-0.25cm}
\label{sec:Task formulation and experimental design}
We sampled Beta distributions to represent varying levels of predicted likelihood and uncertainty, parameterized by mode (\textit{m}) and prior hypothetical sample size (\textit{k}). These relate to the standard Beta shape parameters as $\alpha = m(k-2) + 1$ and $\beta = (1-m)(k-2) + 1$. Note that when $k = 2$, the Beta reduces to $(1,1)$, i.e., a uniform distribution. We selected m values of 0.05, 0.15, 0.25, 0.35, 0.45, 0.55, 0.65, 0.75, 0.85, 0.95 and k values of 50, 150, 250, 350, 450, 550, 650, 750, 850, 950, corresponding to the medians of ten uniform bins over $[0,1]$ and $[0, 1000]$, respectively. All 100 combinations of (m, k) pairs defined a grid of Beta distributions with systematically varying likelihood and uncertainty. To emulate the probabilistic output of an upstream uncertainty-aware model, each ground truth distribution was treated as a generative source. We drew N = 100 samples from each, mimicking the sample-based predictions produced by models that use Monte Carlo Dropout. To see a visualization of how the sampled m and k values jointly control the shape of Beta distributions, please consult Appendix \ref{sec:Beta_shape}. To elicit structured and comparable responses across models, LLMs were constrained to select from a fixed set of NLEs. This framing reduces the task to a text classification problem, enabling systematic comparison across models and experimental conditions. Separate descriptor sets were defined for likelihood and uncertainty, listed in Table~\ref{tab:nle_strings}. Both sets were derived from \citet{Spiegelhalter2011} work on risk communication, and are organized as ordered sequences from least to most probable, and from least to most certain, respectively. This ordering is central to the calibration metric: a well-calibrated model should select descriptors whose position in the sequence reflects the magnitude of the underlying numerical quantity.

\subsection{Prompt design and model selection}\vspace{-0.25cm}
\label{sec:Prompt design and model selection}

Each prompt was instantiated within a specific prediction context, representing a realistic scenario in which an upstream predictive model has produced a probabilistic output that an LLM is asked to communicate to an end-user. Six domain contexts were used: flood prediction, cancer prediction, drug anaphylaxis, gambling, restaurant recommendations, and aurora hikes (Table~\ref{tab:prompt_contexts}). These were selected to vary two properties independently: the level of stakes involved, and whether the predicted event affects a single individual or a population. Individual-level contexts included cancer prediction and drug anaphylaxis, while population-level contexts included flood prediction and aurora hikes. To control for sensitivity to instruction wording, the instruction portion of each prompt was randomly drawn from a set of ten paraphrased variants with equivalent meaning, ensuring that observed variation in model behavior could be attributed to differences in probabilistic inputs or domain contexts rather than surface-level phrasing. Please refer to Appendix \ref{sec:Prompt_example} for an explicit prompt example.

The following models were evaluated: DeepSeekR1-7B, Llama3-8B, Mistral-7B, Olmo3-7B, Qwen3-4B, Qwen3-32B, Gemma3-4B, GPT-4.1-mini and GPT-5.4. Since most of these are small, open source models in the four-to-eight billion parameter range, Qwen-32B was added as additional reference point for larger models. GPT-4.1-mini was added to have a small commercial baseline, and GPT-5.4 was added to have a large, state-of-the-art commercial baseline. To enforce adherence to the required output format, vLLM library \citep{library_2020} was used for guided decoding on open models. For GPT-5.4 and GPT-4.1-mini, guided decoding through vLLM was not available, and was therefore enforced through OpenAI's API.


{\centering\small
\begin{tabular}{ll}
\toprule
\textbf{Likelihood NLEs} & \textbf{Uncertainty NLEs} \\
\midrule
virtually certain        & completely certain \\
very likely              & highly certain \\
likely                   & somewhat certain \\
about as likely as not   & somewhat uncertain \\
unlikely                 & considerably uncertain \\
very unlikely            & highly uncertain \\
exceptionally unlikely   & completely uncertain \\
\bottomrule
\end{tabular}
\captionof{table}{NLE strings for likelihood and uncertainty, organized in an increasing sequence. LLMs where prompted to choose the best NLE for describing the likelihood and the uncertainty in a scenario.}
\label{tab:nle_strings}
\par}


{\centering\small
\begin{tabular}{lcc}
\toprule
\textbf{Domain Context} & \textbf{Stake} & \textbf{Level} \\
\midrule
Flood prediction  & High   & Population \\
Cancer prediction & High   & Individual \\
Drug anaphylaxis  & High   & Individual \\
Gambling odds     & Medium & Individual \\
Gastronomy        & Low    & Individual \\
Aurora hikes  & Low    & Population \\
\bottomrule
\end{tabular}
\captionof{table}{To test for model robustness, different domain contexts were used to modulate the level of stakes in a situation, across individual and population level events.}
\label{tab:prompt_contexts}
\par}

\subsection{Evaluation metrics}\vspace{-0.25cm}

\subsubsection{Model consistency} Consistency Score \citep{Kolena2024} was calculated from the ten repetitions performed for each scenario, measuring how often the model selected the same descriptor across identical inputs. Consistency score ranged between [0, 1], where higher values indicate greater consistency. It was calculated for each predictive scenario at each temperature setting, averaging only across the ten inference repetitions. To ensure that consistency was measured across truly identical prompts, the instruction paraphrase was held constant across all ten repetitions within a given scenario. Variation in instruction wording occurred only between scenarios, not within the repetitions used to compute consistency scores. All consistency scores include 95\% confidence intervals. The formula for Consistency Score is located in Appendix \ref{sec:consistency}.

\subsubsection{Model calibration} We evaluated calibration by measuring how often a model's NLE choices contradicted the ordering implied by its own usage across predictive scenarios. Therefore, if a model used \textit{``likely''} for a lower mode value but \textit{``unlikely''} for a higher one, this was counted as a violation, since the model's own behavior elsewhere implied that \textit{``likely''} should correspond to higher values than \textit{``unlikely''}. On this assumption, we calculated a modified version of the Jonckheere-Terpstra (JT) trend test, which tests whether values increase consistently across groups arranged in a known order, by counting how often a value in a higher-ranked group actually exceeds a value in a lower-ranked group and then summing those counts across all group pairs \citep{Ali2015}. This test was applied separately for likelihood and uncertainty. For likelihood, each NLE (e.g., ``unlikely'', ``likely'', ``very likely'') represented a group, and the values being compared were the distribution modes (\textit{m}) associated with each descriptor. For uncertainty, each NLE (e.g., ``somewhat uncertain'', ``highly certain'') represented a group, and the values compared were the hypothetical prior sample sizes (\textit{k}). 

A penalization was added for LLMs utilizing fewer groups of NLEs than the maximum available groups for a task. This scenario could happen when an LLM did not choose an NLE string a single time for a given experiment. We penalized such cases in two ways, logarithmic and power penalties, where parameter $\beta$ controls the strength of the penalty. To ensure it wasn't the penalty that dominated the score over the violation of own past criteria, beta parameter was set at $\beta = 0.5$. The result was a metric that ranged between $[0,1]$, where the higher the value, the better. The resulting calibration score was calculated at the most granular level of the experimental design: per LLM, domain context, temperature, and repetition. For likelihood calibration, scores were computed across distribution modes. For uncertainty calibration, they were computed across hypothetical prior sample sizes. Summarized averages and 95\% confidence intervals were then produced from these disaggregated calculations. The formula for Calibration Score is in Appendix \ref{sec:calibration}.

\subsubsection{Task correctness (pass@k)}
LLMs may produce outputs not conforming to the guided decoding format at higher temperatures, which must be excluded when calculating scores. These exclusions could inflate scores, since fewer observations per group reduce the probability of violating past criteria (artificially increasing calibration) and of non-concordant pairs (artificially increasing consistency). To account for this, pass@k was calculated for both scores, measuring the probability that a model produces a correctly formatted output within k=5 attempts. Lower pass@k values signal that corresponding calibration and consistency scores should be interpreted with caution. All averaged scores include 95\% confidence intervals. The pass@k formula is located in Appendix \ref{sec:passatk_score}.

\subsubsection{Experiment ablations}

Every combination of predictive scenario and repetition was additionally iterated across ten LLM temperature values, which were: 0.0, 0.1, 0.2, 0.3, 0.4, 0.5, 0.6, 0.7, 0.8, 0.9, to study the effect of temperature on model performance. To evaluate whether LLM performance improves when models are not required to derive statistical properties from raw samples, we conducted an ablation in which the precomputed mode (\textit{m}) and hypothetical prior sample size (\textit{k}) of each distribution were provided directly in the prompt. All other experimental conditions remained identical. This ablation removes the need for models to infer the statistical properties of the distribution before selecting an NLE. Because the provided values were derived from the ground truth parameters used to generate each distribution, this ablation represents an idealized scenario in which statistical inference is performed without error. All codes and data necessary for the reproducibility of this method are available in the (anonymized) Github at Appendix \ref{sec:git} and on the front-page link.

\subsubsection{Metric robustness ablations}

To assess the robustness of the calibration metric and the generality of our findings, we conducted six additional ablations beyond our standard experiments. First, we re-scored every model's descriptor choices against externally validated probability ranges drawn from \citet{Spiegelhalter2011} and the \citet{IPCC2018}, computing mean signed and absolute error against each descriptor's reference midpoint. Second, we performed a penalty-parameter sensitivity analysis by re-computing calibration under the power-penalty variant for $\beta \in \{0, 0.25, 0.5, 0.75, 1.0\}$ on both tasks, where $\beta = 0$ corresponds to the absence of any vocabulary-coverage penalty. Third, we ran a descriptor-scale ablation at 5 descriptors across eight of the nine manuscript models (Llama-3-8B excluded). The likelihood set was anchored to Hashim (2024)'s empirically validated terms, while the uncertainty set was constructed by removing the two most semantically adjacent terms ("somewhat certain" / "somewhat uncertain") from the original. Fourth, we re-scored calibration against empirical sample means and variances derived from the N = 100 samples, rather than against the population parameters m and k. Fifth, motivated by our variance-based working definition of predictive uncertainty, we re-scored uncertainty calibration against the Beta variance, with standard deviation, entropy, and central credible-interval width tested as alternative spread measures. Finally, we decomposed the domain-context effect along two principled design dimensions: stake-level (high, medium and low) and event-scope (individual and population), by re-aggregating per-context calibration scores along each axis.

\end{multicols}
\section{Results}\vspace{-0.25cm}
\label{sec:Results}

\begin{multicols}{2}

\begin{figure*}[t]
  \begin{center}
  \includegraphics[width=1\linewidth]{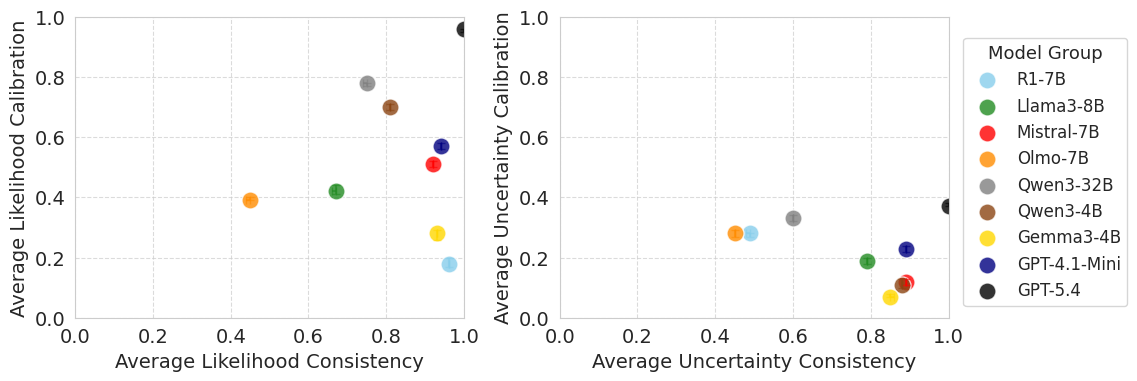}
  \end{center}
  \caption{Scatterplot of LLM calibration and consistency for explaining the likelihood of an event occurring. 95\% confidence intervals are plotted but may not be visible at this scale due to their small magnitude. Appendix \ref{sec:mean_numbers} shows the mean calibration scores used to generate this plot.}
  \label{fig:summary_scatter}
\end{figure*}

Overall, LLMs were found to be consistent but miscalibrated for describing the likelihood of an event, and particularly miscalibrated for describing uncertainty (Figure \ref{fig:summary_scatter}, Appendix \ref{sec:mean_numbers}). We found that domain context alone can influence calibration and consistency, and that increasing temperature has a trade-off, slightly increasing calibration but considerably decreasing consistency. Task correctness was extremely high across LLMs, with most architectures averaging a pass@5 of 1.0 across scenarios (Appendix \ref{sec:passatk_table}).

\subsection{LLM consistency}\vspace{-0.25cm}

Consistency for likelihood was generally high, with all architectures achieving mean scores above 0.8, except Qwen3-32B, Llama3-8B and Olmo3-7B. Similar trends were observed for uncertainty tasks, except for R1-7B decreasing considerably to 0.49, and Llama3-8B improving to 0.79 (Figure \ref{fig:summary_scatter}). GPT-5.4 performed considerably better than all other architectures, achieving a perfect mean consistency score of 1.0 both for likelihood and uncertainty tasks. Domain context was found to affect consistency in both likelihood and uncertainty tasks, with some architectures more prone to effect than others. Not one context was found to uniformly improve or worsen consistency (Appendix \ref{sec:dissag_con}).

\subsection{LLM calibration}\vspace{-0.25cm}

\begin{figure*}[t]
  \begin{center}
  \includegraphics[width=1\linewidth]{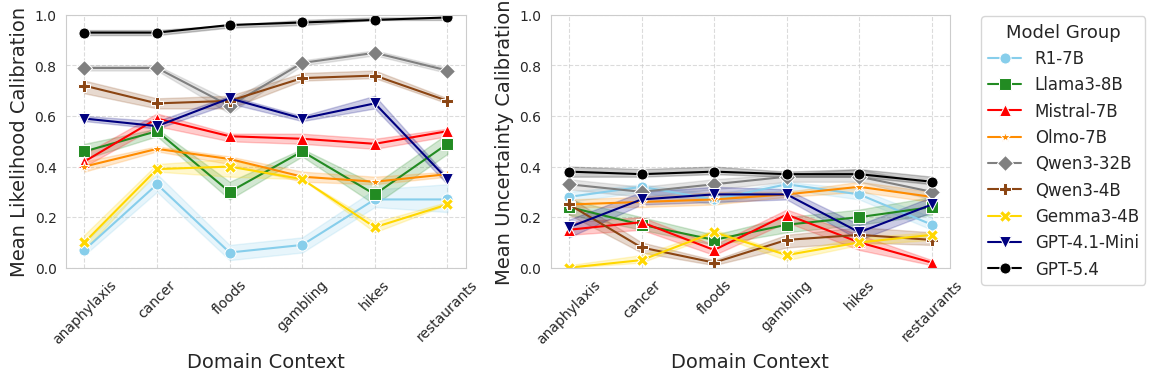}
  \end{center}
  \caption{Model calibration across domain contexts for describing likelihood and uncertainty. 95\% confidence intervals are plotted but may not be visible due to their small magnitude.}
  \label{fig:llm_context_cali}
\end{figure*}

Except for GPT-5.4, which achieved a mean likelihood calibration of 0.96, and both Qwen3 variants (32B and 4B) with 0.78 and 0.70, respectively, LLMs were generally found to be miscalibrated (Figure~\ref{fig:summary_scatter}) for likelihood tasks.  For uncertainty tasks, scores were considerably lower across all models, including GPT-5.4: no architecture achieved scores above 0.40, and some (Gemma3-4B) scored as low as 0.07 (Figure~\ref{fig:summary_scatter}). Domain context was also found to meaningfully affect calibration across architectures. However, no consistent patterns emerged across specific domains or architectures (Figure~\ref{fig:llm_context_cali}). As with previous trends, GPT-5.4 deviated from the norm, remaining generally unaffected by changes in domain context. There was no observable effect of hypothetical prior sample size (\textit{k}) on likelihood calibration, or of distribution mode (\textit{m}) on uncertainty calibration (Appendix~\ref{sec:dissag_cali}).

\subsection{Vocabulary mappings}\vspace{-0.25cm}

Most models failed to select all available NLE descriptors (Appendix \ref{sec:vocab_map} \& \ref{sec:dissag_vocab_map}). For likelihood tasks, only GPT-5.4, Qwen3 (4B and 32B), and Olmo3-7B selected every likelihood descriptor at least once. Among these, only the best calibrated architectures consistently shifted descriptor choices as distribution mode increased, suggesting an internal mapping between numerical likelihood and descriptor hierarchy. GPT-5.4 exhibited the sharpest mapping: as mode (\textit{m}) increased, the model assigned a single deterministic descriptor per mode value, producing a clean stepwise correspondence between likelihood and verbal descriptors, notably without the mode being explicitly provided. For uncertainty tasks, no model selected all available descriptors. All architectures defaulted to a narrow set of moderate descriptors (such as ``somewhat certain'' or ``somewhat uncertain'') regardless of underlying mathematical uncertainty, while more decisive descriptors were rarely selected.

\subsection{Effect of temperature and precomputed data on model performance}\vspace{-0.25cm}

Increasing temperature significantly decreased consistency for both likelihood and uncertainty tasks  and slightly improved calibration, across most architectures, save for GPT-5.4, whose metrics remained stable across temperatures (Appendices \ref{sec:consistency_temp}, \ref{sec:cali_temp}). Providing precomputed ground-truth parameters, that is, the distribution mode (m) and hypothetical prior sample size (k), did not improve calibration or consistency for any model or task (Appendix \ref{sec:precomputed_ablation}). Performance slightly decreased for several smaller architectures, except the Qwen3 family, which remained largely unaffected, while GPT-5.4 maintained identical performance. One notable effect was a reduction in calibration score variability across domain contexts, suggesting that explicit numerical summaries may anchor model behavior and reduce sensitivity to contextual framing (Appendix \ref{sec:precomputed_context}).

\subsection{Metric robustness ablations}

All six ablations support the manuscript's qualitative results, with two refinements to specific magnitude claims. External validation against the expert elicited likelihood descriptors \citep{Spiegelhalter2011, IPCC2018} confirmed that GPT-5.4's mappings are not merely monotonic but track expert consensus closely (mean signed error $-0.01$, mean absolute error 0.08), while the other seven models showed substantially larger absolute errors (0.24–0.41). Signed errors were negative across all nine architectures, indicating a systematic tendency to overstate likelihood. 

The penalty-sensitivity sweep for the power-penalty variation of our calibration metric showed GPT-5.4 ranking first at every $\beta$ on both tasks, including at $\beta = 0$, ruling out the vocabulary-coverage penalty as the source of its lead. Meanwhile, mid-tier LLM rankings shifted with $\beta$ as intended, with GPT-4.1-mini and Gemma3-4B declining at higher $\beta$, as they relied on a narrower descriptor set.

The ablation done on 5-descriptor scale showed similar trends as our standard 7-descriptor scale experiment. Likelihood calibration scores tended to compress towards the middle-range, with GPT-5.4 decreasing from 0.96 to 0.79, and Qwen3-32B decreasing from 0.78 to 0.67, while R1-7B rose from 0.18 to 0.39. In contrast, uncertainty calibration scores decreased even further, with no model exceeding 0.24 under this ablation. 

Scoring against empirical sample statistics rather than population parameters shifted scores by less than $\pm$ 0.005 for eight of the nine models, with GPT-5.4 having largest deviation (likelihood 0.96 to 0.98; uncertainty 0.37 to 0.39), confirming that the calibration assessment is not an artifact of access to hidden population parameters. Variance-based uncertainty scoring raised scores meaningfully for top-tier models (GPT-5.4 0.37 to 0.57; Qwen3-32B 0.33 to 0.54; GPT-4.1-mini 0.23 to 0.46; Olmo 0.28 to 0.42), with four architectures exceeding the 0.40 threshold under this measure. Standard deviation, entropy, and credible-interval width yielded near-identical scores. Now, our previous statement of "no architecture above 0.40" statement is specifically in relation to concentration-based scoring, rather than to predictive uncertainty in general.

Finally, stake and scope decomposition produced no systematic directional effect along either dimension across the nine models, corroborating the "no consistent patterns" claim, though the unbalanced six-context design (in particular the absence of a medium-population sized event cell) precludes a confirmatory factorial test. All data generated for these additional ablations can be found in the chapter Thesis Supplementary Material.

\end{multicols}
\section{Discussion}\vspace{-0.25cm}

\begin{multicols}{2}

Our study demonstrated that LLMs are generally consistent but miscalibrated when describing the likelihood and uncertainty of predicted events in natural language, with domain context and temperature directly affecting performance. These results suggest that most modern LLM architectures lack a robust internal mapping of likelihood and uncertainty, and consequently struggle to express them in a manner calibrated to the underlying probabilistic information.

GPT-5.4's performance on likelihood tasks deviates from the general pattern. Its near perfect calibration and consistency scores suggest the model has internalized a robust, deterministic mapping between the statistical mode of a distribution and verbal likelihood descriptors. Whether this mapping aligns with expert or lay-people interpretations of risk remains to be studied. GPT-5.4 consistently chose one NLE descriptor per mode value, indicating sharply defined and stable internal hedging boundaries across experimental conditions. The diminished effect of domain context on its calibration suggests that likelihood is treated as a context-invariant numerical property rather than a domain sensitive judgment. These characteristics align with what has been described as agentic capability in frontier models: the capacity to execute a well defined subtask reliably and autonomously, without requiring external scaffolding or domain specific adaptation. However, this competence did not transfer to uncertainty tasks, where GPT-5.4 obtained a mean score of 0.37 despite perfect consistency. Like all other evaluated models, GPT-5.4 defaulted to a narrow band of moderate descriptors regardless of mathematical uncertainty, suggesting that even frontier scale models have not developed reliable internal hedges for predictive uncertainty.

Increasing temperature offered a tradeoff, improving calibration but worsening consistency. We believe this is driven by the effect of temperature on token selection \citep{renze-2024-effect}. Higher temperatures increase the probability of less frequent tokens being sampled, which had two consequences. First, models were more likely to shift to a different descriptor as the underlying distribution changed and to draw from a wider vocabulary, both of which improve calibration under our definition. Second, the same mechanism made it more likely for a model to select different descriptors for identical scenarios across repetitions, reducing consistency. This is consistent with findings from \citet{harsha_agarwal_lakkaraju_2023}, who used temperature as a probing strategy to measure uncertainty in LLM explanations. In their work, higher temperatures produced greater variation in generated explanations for identical inputs, interpreted as lower explanation reliability. Coupled with our results, these findings suggest that temperature trades off stability for expressiveness.

An initial hypothesis for poor calibration, particularly for uncertainty, was that numerical inputs were presented as raw lists of sampled values, requiring models to derive summary statistics before selecting a descriptor. However, our precomputed ablation, in which the distribution mode and hypothetical prior sample size were stated explicitly, did not improve performance. This suggests that the difficulty lies not in statistical inference over raw samples but in mapping known numerical quantities to calibrated verbal descriptors. The bottleneck therefore appears to reside in the verbalization step rather than in numerical reasoning. The ablation did reduce calibration score variability across domain contexts, suggesting that explicit numerical anchors may partially attenuate the influence of contextual framing on descriptor selection. Another possible contributing factor is the training data itself. \citet{LinHiltonEvans2022} demonstrated that LLMs can inherit and reproduce falsehoods present in human generated text, suggesting that if human communicators are themselves poorly calibrated when describing probabilistic information \citep{Hashim2024, Gigerenzer2007}, then models trained on such language may internalize those same miscalibrations.

Our findings suggest that increasing model scale alone may not suffice to produce an emergent understanding of likelihood and uncertainty, though this should be interpreted cautiously given that only one model at roughly 32 billion parameters was included (Qwen3). What our results do show is that strong performance did not arise uniformly across architectures, and that broader experiments across additional mid-to-large scale models are needed before drawing conclusions about scale-driven emergence.

Models currently default to hesitant descriptors rather than clear statements calibrated to the underlying probabilistic information, particularly for uncertainty tasks. This is consistent with a known tendency in LLMs to default to cautious, noncommittal language rather than making precise statements that could be clearly wrong \citep{Sharma2023}. Such ambivalence may be especially problematic in high stakes settings such as floods, cancer, or anaphylaxis, where vague language could affect important downstream decisions. However, some of this caution may reflect appropriate restraint rather than deficiency; our calibration score penalizes models for not utilizing all available descriptors, which may disadvantage models that deliberately avoid extreme language in sensitive contexts. This interpretation is further supported by the observed effect of domain context, where models may use more cautious language in medical and flood contexts than in entertainment contexts as a form of learned sensitivity. Disentangling genuine miscalibration from socially appropriate hedging remains an open question when interpreting domain level differences.

Risk communication is itself difficult even for humans. Translating likelihood and uncertainty into clear, faithful, and interpretable language requires statistical literacy, domain sensitivity, and careful framing, as evidenced by well documented miscalibration in how experts and lay audiences use probabilistic language \citep{Hashim2024}. Considering this, we argue that LLMs remain promising tools, not because they understand probabilistic concepts reliably, but because of their capacity to produce fluent, accessible natural language suited for the verbalization stage of risk communication. Moreover, given the rapid pace at which LLM performance has improved across benchmarks \citep{zhao2026surveylargelanguagemodels}, there is reasonable expectation that the calibration and consistency gaps identified here may narrow as architectures and training methods advance.

Our study has limitations. The main one is that our method relied on a discrete formulation of risk communication through guided decoding and predetermined NLE descriptors. This restriction may shape model behavior and could affect calibration by forcing judgments into a coarse categorical space. This was, however, a necessary decision, as unconstrained outputs would have made systematic comparison across models and conditions considerably more difficult. Another limitation is that real world LLM chatbots are almost exclusively used for long form outputs. Important dimensions of risk communication therefore remain outside our scope, including whether models suggest appropriate downstream actions, contextualize consequences, acknowledge alternative outcomes, or communicate tradeoffs useful to end users. Evaluating LLM performance on these dimensions represents a necessary direction for future work. Additionally, our experiment used synthetic Beta distributed predictions rather than real model outputs, which improves control but may not fully reflect real world scenarios. Finally, all experiments were conducted in English, meaning the observed patterns could be influenced by properties specific to the English language. Without cross linguistic comparison, it is not possible to disentangle the effect of language on model performance.

\end{multicols}
\section{Conclusion}\vspace{-0.25cm}

\begin{multicols}{2}

To conclude, our study provided a systematic evaluation of LLMs for explaining predictive likelihood and uncertainty in natural language, and found that they are generally consistent but miscalibrated, with performance worsening particularly for uncertainty tasks. These findings reinforce that risk communication is itself a difficult and specialized task, even for humans. Together, our results suggest that current LLMs do not yet possess a sufficiently stable internal mapping between probabilistic information and NLEs. While LLMs remain promising tools for producing fluent and accessible language, pairing them with upstream statistical methods alone does not resolve the calibration deficits identified here, underscoring the need for targeted interventions at the verbalization stage itself.

\end{multicols}
\section{Acknowledgements}\vspace{-0.25cm}

\begin{multicols}{2}

This work was funded and supported by the Canadian Institutes of Health Research (CIHR) Project Grant [PJT-206007, PJT-173272], IVADO and the Canada First Research Excellence Fund Regroupement R3 Grant, and the Network for Canadian Oral Health Research (NCOHR) Seed Grant. S.M. is supported by a Fonds de recherche du Québec – Santé (FRQS) Salary Award. Computational infrastructure was supported by the Canada Foundation for Innovation (CFI) John R. Evans Leaders Fund (JELF) [44548].

\end{multicols}

\newpage
\appendix
\section{Appendix}

\subsection{Effect of parameters m and k on the shape of Beta distributions}
\label{sec:Beta_shape}

\begin{figure}[H]
  \begin{center}
  \includegraphics[width=0.55\linewidth]{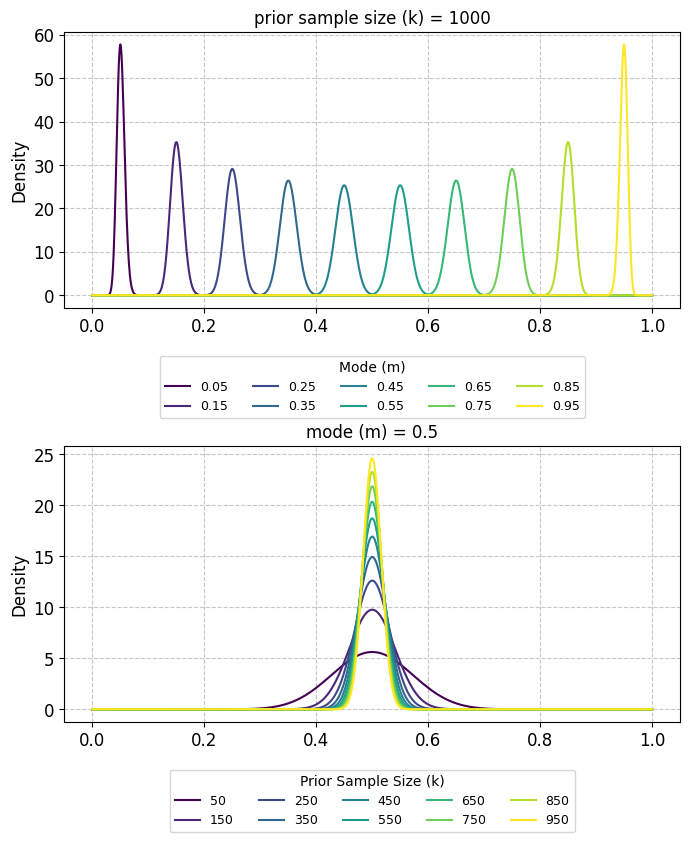}
  \end{center}
  \caption{Shape change of beta distributions as mode (\textit{m}) and hypothetical prior sample size (\textit{k}) are changed. Both parameters were used to modulate the likelihood and the uncertainty present in a distribution. The higher the k parameter, the higher the certainty. The higher the m parameter, the higher the likelihood.}
  \label{fig:sampling_probs}
\end{figure}

\subsection{Explicit prompt example}
\label{sec:Prompt_example}

\begin{tcolorbox}
"A model was developed to predict the probability of catastrophic floods occurring in a region next year. Analyze these values carefully, and create a summary of the overall prediction. Make sure that you summarize the likelihood of the event (predicted risk) and the predictive uncertainty (uncertainty in the predicted risk). To verbalize the likelihood of the event, you must use one of the following list of words: 'virtually certain', 'very likely', 'likely', 'about as likely as not', 'unlikely', 'very unlikely', 'exceptionally unlikely.' To verbalize the predictive uncertainty, you must use one of the following list of words: 'completely certain', 'highly certain', 'somewhat certain', 'somewhat uncertain', 'considerably uncertain', 'highly uncertain', 'completely uncertain'. You must choose only one word from each of the two lists per given set of risk values. Here are the risk estimates: 0.08, 0.06, 0.12, (...), 0.08. Only answer back with the chosen words."
\end{tcolorbox}

\subsection{Domain context variations}
\label{sec:domain_contexts}
\begin{enumerate}
    \item{\textbf{Flood prediction}}: "A model was developed to predict the probability of catastrophic floods occurring in a region next year..."
    \item{\textbf{Cancer prediction}}: "You are a doctor who has consulted an AI model on the probability of a patient having cancer given their clinical characteristics, to help determine whether to initiate chemotherapy..."
    \item{\textbf{Drug anaphylaxis}}: "You are evaluating the possibility of a patient developing anaphylactic shock when prescribed medication for severe pain..."
    \item{\textbf{Gambling}}: "You are a gambler evaluating whether a certain bet is worth taking. For this, you used an AI agent that runs simulations and predicts the probabilities of the odds being in your favor..."
    \item{\textbf{Restaurant recommendations}}: "You are in charge of recommending the best current restaurants in the San Francisco Bay Area to customers seeking reservations..."
    \item{\textbf{Aurora Hikes}}: "You are communicating the probability of Aurora Borealis sightings in a tourist town in the mountains. When there are high chances of Aurora sightings, many tourists are expected to make night hikes..."
\end{enumerate}

\subsection{Instruction paraphrases}
\label{sec:instructions}

\begin{enumerate}
    \item{\textbf{Instruction 1:} To express how likely the event is, you may choose only one term from the following list: (...). To express how uncertain the event is, you may choose only one term from the following list: (...)}
    \item{\textbf{Instruction 2:} To verbalize the likelihood of the event, you must use one of the following list of words: (...). To express how uncertain the event is, you may choose only one term from the following list: (...)}
    \item{\textbf{Instruction 3:} Please indicate the probability of the event using a single string from the following list: (...). Select a single string from the list below to express the uncertainty that the event will occur: (...)}
    \item{\textbf{Instruction 4:} Please indicate the probability of the event using a single string from the following list: (...). To describe the uncertainty of the event, restrict your choice to one string from the list provided: (...)}
    \item{\textbf{Instruction 5:} Please indicate the probability of the event using a single string from the following list: (...). Please indicate the uncertainty of the event using a single string from the following list: (...)}
    \item{\textbf{Instruction 6:} To verbalize the likelihood of the event, you must use one of the following list of words: (...). Use only one of the strings in this list to convey the uncertainty of the event: (...)}
    \item{\textbf{Instruction 7:} To describe the likelihood of the event, restrict your choice to one string from the list provided: (...). Select a single string from the list below to express the uncertainty that the event will occur: (...)}
    \item{\textbf{Instruction 8:} To verbalize the likelihood of the event, you must use one of the following list of words:(...). To describe the uncertainty of the event, restrict your choice to one string from the list provided: (...)}
    \item{\textbf{Instruction 9:} To describe the likelihood of the event, restrict your choice to one string from the list provided: (...). To describe the uncertainty of the event, restrict your choice to one string from the list provided: (...)}
    \item{\textbf{Instruction 10:} Use only one of the strings in this list to convey the likelihood of the event: (...). Select a single string from the list below to express the uncertainty that the event will occur: (...)}
\end{enumerate}

\subsection{Evaluation metric formulas}
\subsubsection{Consistency score}
\label{sec:consistency}

  \begin{equation}
\label{eq:consistency_compact}
CS = \frac{N_{\text{consistent pairs}}}{N_{\text{total pairs}}}
\end{equation}

\subsubsection{Calibration score}
\label{sec:calibration}

Assuming the correct ordering of a sequence groups as $G_1 < G_2 < \dots < G_k$, the Jonckheere--Terpstra statistic was defined as:

\begin{equation}
\label{eq:jt_stat}
J = \sum_{i < j}
    \sum_{x \in G_i}
    \sum_{y \in G_j}
    \phi(x,y)
\end{equation}

where:

\begin{equation}
\label{eq:phi}
\phi(x,y) =
\begin{cases}
1   & \text{if } y > x \\
0.5 & \text{if } y = x \\
0   & \text{if } y < x
\end{cases}
\end{equation}
This way, the maximum possible value of the statistic is:

\begin{equation}
\label{eq:jmax}
J_{\max} = \sum_{i < j} n_i n_j
\end{equation}

where $n_i$ denotes the number of samples in group $G_i$. Further, a normalized version of the score was expressed as:

\begin{equation}
\label{eq:jtnorm}
JT_{\text{normalized}} = \frac{J}{J_{\max}}
\end{equation}

which was later penalized through logarithmic and power penalties, such that:

\begin{equation}
\label{eq:jt_log}
JT_{\text{log-penalty}} =
JT_{\text{norm}} \times
\frac{\log(k)}{\log(k_{\max})}
\end{equation}

\begin{equation}
\label{eq:jt_power}
JT_{\text{power-penalty}} =
JT_{\text{norm}} \times
\left(\frac{k}{k_{\max}}\right)^{\beta}
\end{equation}

\subsubsection{Pass@k}
\label{sec:passatk_score}

\begin{equation}
\text{pass@}k = 1 - \frac{\binom{n - c}{k}}{\binom{n}{k}}
\label{eq:pass_at_k}
\end{equation}

where $n$ is the total number of inference attempts, $c$ is the number of correctly formatted outputs, and $k$ is the allowed attempt budget.

\subsection{Anonymized Github Link}
\label{sec:git}

All necessary data, codes and instructions to replicate the results in this study can be found inside an anonymized GitHub, link \href{https://anonymous.4open.science/r/COLM_submission-0B47/README.md}{here}.

\subsection{Mean calibration scores for likelihood and uncertainty tasks per LLM.}
\label{sec:mean_numbers}

\begin{table}[ht]
\centering
\caption[Mean calibration and consistency scores per LLM]{Mean calibration and consistency scores for likelihood and uncertainty tasks, under logarithmic penalty, by model group, along with 95\% confidence intervals shown in brackets.}

\label{tab:mean_cali_averages}
\vspace{1ex}
\resizebox{\textwidth}{!}{%
{\small
\begin{tabular}{lcccc}
\toprule
\textbf{Model group} & \textbf{Likelihood calibration} & \textbf{Uncertainty calibration} & \textbf{Likelihood consistency} & \textbf{Uncertainty consistency} \\
\midrule
R1 7B        & 0.18 [0.17, 0.20] & 0.28 [0.27, 0.28] & 0.96 [0.96, 0.96] & 0.49 [0.48, 0.49] \\
Llama3 8B    & 0.42 [0.41, 0.44] & 0.19 [0.18, 0.20] & 0.67 [0.66, 0.67] & 0.79 [0.79, 0.79] \\
Mistral 7B   & 0.51 [0.50, 0.52] & 0.12 [0.11, 0.13] & 0.92 [0.92, 0.92] & 0.89 [0.89, 0.89] \\
Olmo3 7B     & 0.39 [0.39, 0.40] & 0.28 [0.27, 0.29] & 0.45 [0.44, 0.45] & 0.45 [0.45, 0.45] \\
Qwen3 32B    & 0.78 [0.77, 0.78] & 0.33 [0.32, 0.34] & 0.75 [0.74, 0.75] & 0.60 [0.60, 0.60] \\
Qwen3 4B     & 0.70 [0.69, 0.71] & 0.11 [0.10, 0.13] & 0.81 [0.81, 0.81] & 0.88 [0.88, 0.88] \\
Gemma3 4B    & 0.28 [0.26, 0.29] & 0.07 [0.07, 0.08] & 0.93 [0.93, 0.93] & 0.85 [0.85, 0.86] \\
GPT 4.1 mini & 0.57 [0.56, 0.58] & 0.23 [0.22, 0.24] & 0.94 [0.94, 0.94] & 0.89 [0.89, 0.89] \\
GPT 5.4      & 0.96 [0.95, 0.96] & 0.37 [0.37, 0.38] & 1.00 [1.00, 1.00] & 1.00 [1.00, 1.00] \\
\bottomrule
\end{tabular}
}%
}
\end{table}

\subsection{Effect of temperature on LLM Consistency for describing likelihood and uncertainty.}
\label{sec:consistency_temp}

\begin{figure}[H]
  \begin{center}
  \includegraphics[width=1\linewidth]{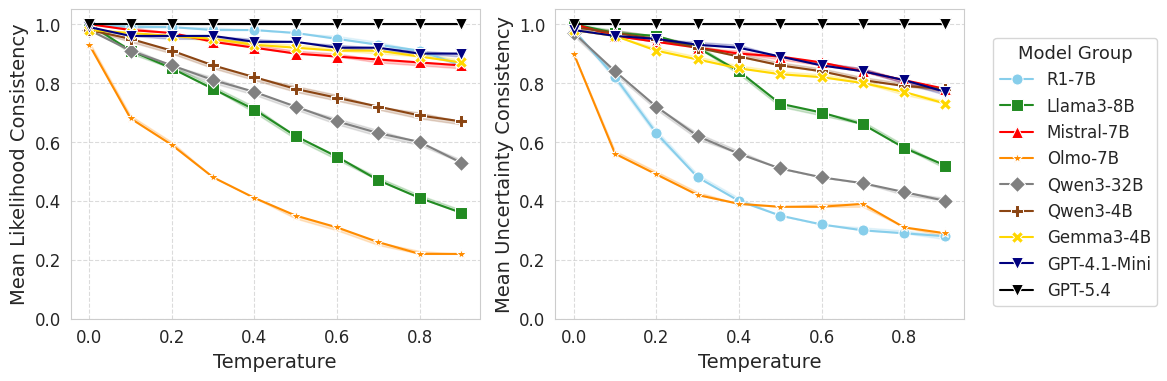}
  \end{center}
  \caption{LLM consistency for explaining the likelihood and the uncertainty of an event occurring, as temperature increases. 95\% confidence intervals are plotted but may not be visible at this scale due to their small magnitude.}
  \label{fig:LLM_consistency_temp}
\end{figure}

\subsection{Effect of temperature on LLM Calibration for describing likelihood and uncertainty.}
\label{sec:cali_temp}

\begin{figure}[H]
  \begin{center}
  \includegraphics[width=1\linewidth]{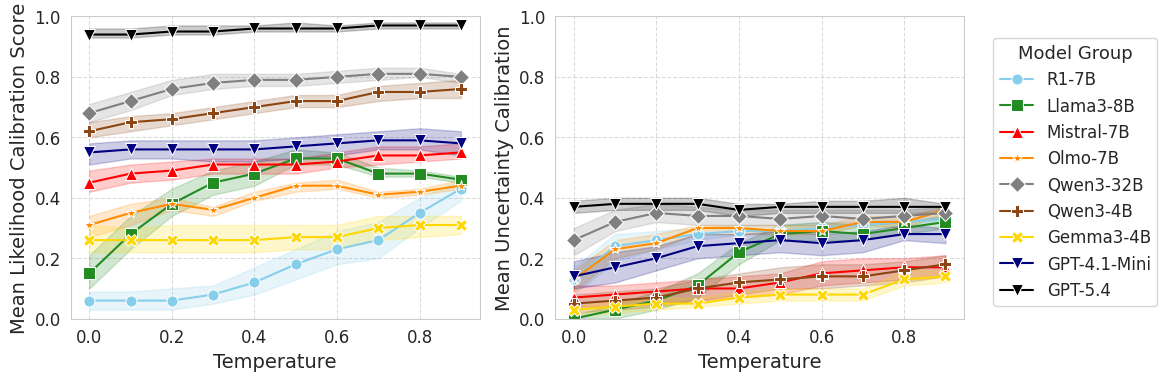}
  \end{center}
  \caption{LLM calibration for explaining the likelihood of an event occurring, as temperature increases, along 95\% confidence intervals.}
  \label{fig:cali_temp}
\end{figure}

\subsection{Effect of precomputed data on LLM Calibration and Consistency for describing likelihood and uncertainty.}
\label{sec:precomputed_ablation}

\begin{figure}[H]
  \begin{center}
  \includegraphics[width=1\linewidth]{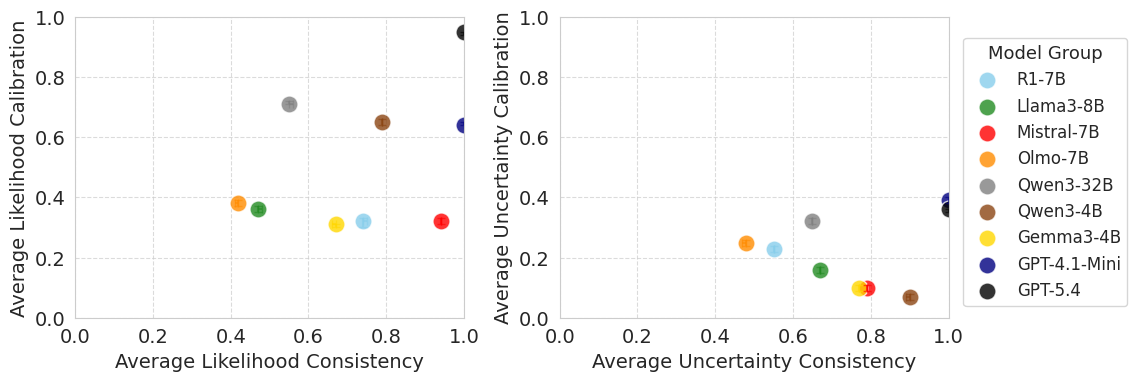}
  \end{center}
  \caption{Scatterplot of LLM calibration and consistency for explaining the likelihood of an event occurring with ground truths explicitly stated in the prompt. 95\% confidence intervals are plotted but may not be visible at this scale due to their small magnitude.}
  \label{fig:gt_summary_scatter}
\end{figure}

\subsection{Effect of precomputed data on effect of domain context, for likelihood and uncertainty tasks, per LLM.}
\label{sec:precomputed_context}

\begin{figure}[H]
  \begin{center}
  \includegraphics[width=1\linewidth]{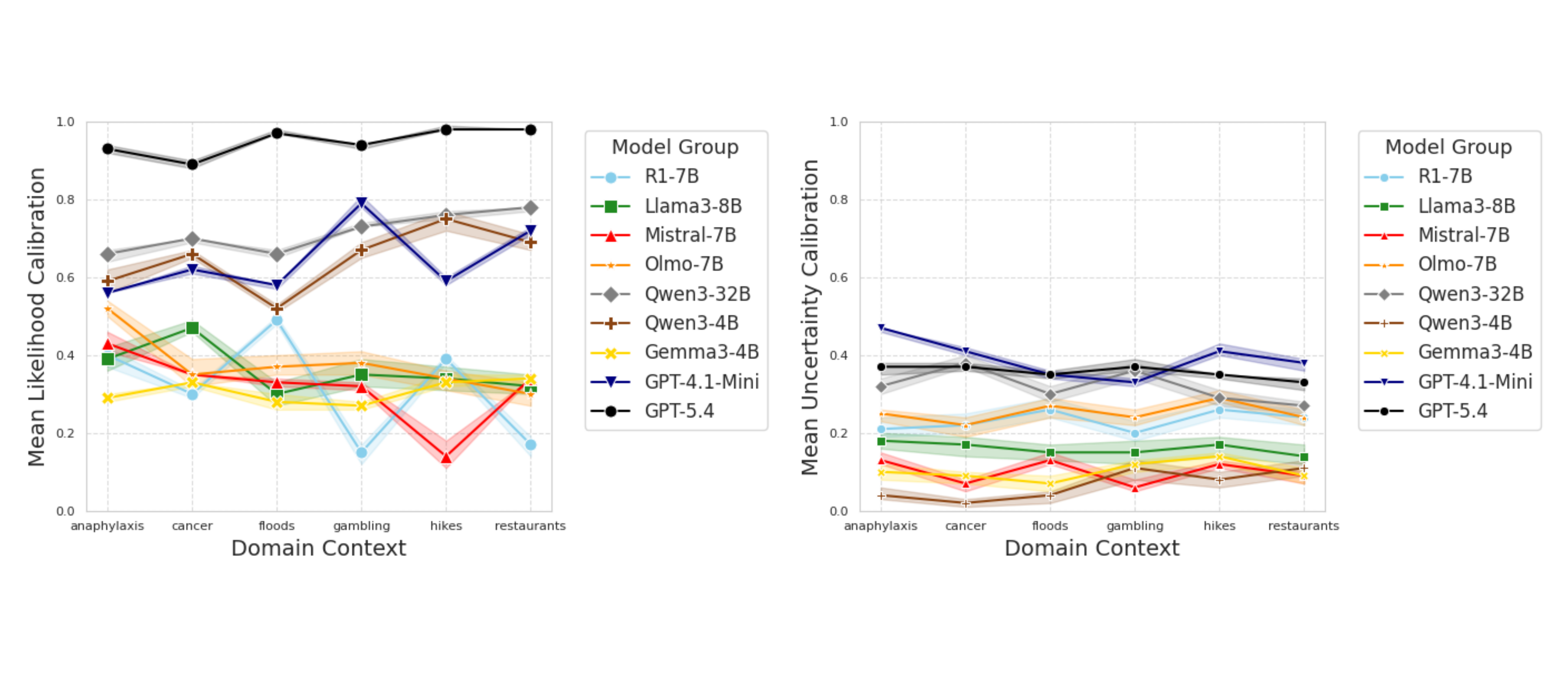}
  \end{center}
  \caption{Scatterplot of LLM calibration, across domain contexts, for explaining the likelihood and the uncertainty of an event with ground truths explicitly stated in the prompt. 95\% confidence intervals are plotted but may not be visible at this scale due to their small magnitude.}
  \label{fig:gt_context}
\end{figure}

\subsection{Mean pass@5 values by LLM (task-correctness)}
\label{sec:passatk_table}

\begin{table}[H]
\begin{center}
\begin{tabular}{lcc}
\toprule
\textbf{Model} & \textbf{Likelihood} & \textbf{Uncertainty} \\
\midrule
R1-7B & 1.00 & 1.00 \\
Llama3.1-8B & 0.99 & 0.99 \\
Mistral-7B & 0.99 & 0.99 \\
Olmo3-7B & 0.99 & 0.99 \\
Qwen3-32B & 1.00 & 1.00 \\
Qwen3-4B & 1.00 & 1.00 \\
Gemma3-4B & 1.00 & 1.00 \\
GPT-4.1-mini & 1.00 & 1.00 \\
GPT-5.4 & 1.00 & 1.00 \\
\bottomrule
\end{tabular}
\end{center}
\caption{Mean pass@5 scores for likelihood and uncertainty tasks across all evaluated models. Most architectures achieved near-perfect task correctness.}
\label{tab:pass_at_k}
\end{table}

\subsection{Disaggregated calibration scores (likelihood and uncertainty) per LLM, across experiment temperatures}
\label{sec:dissag_cali}

\begin{figure}[H]
  \begin{center}
  \includegraphics[width=\linewidth]{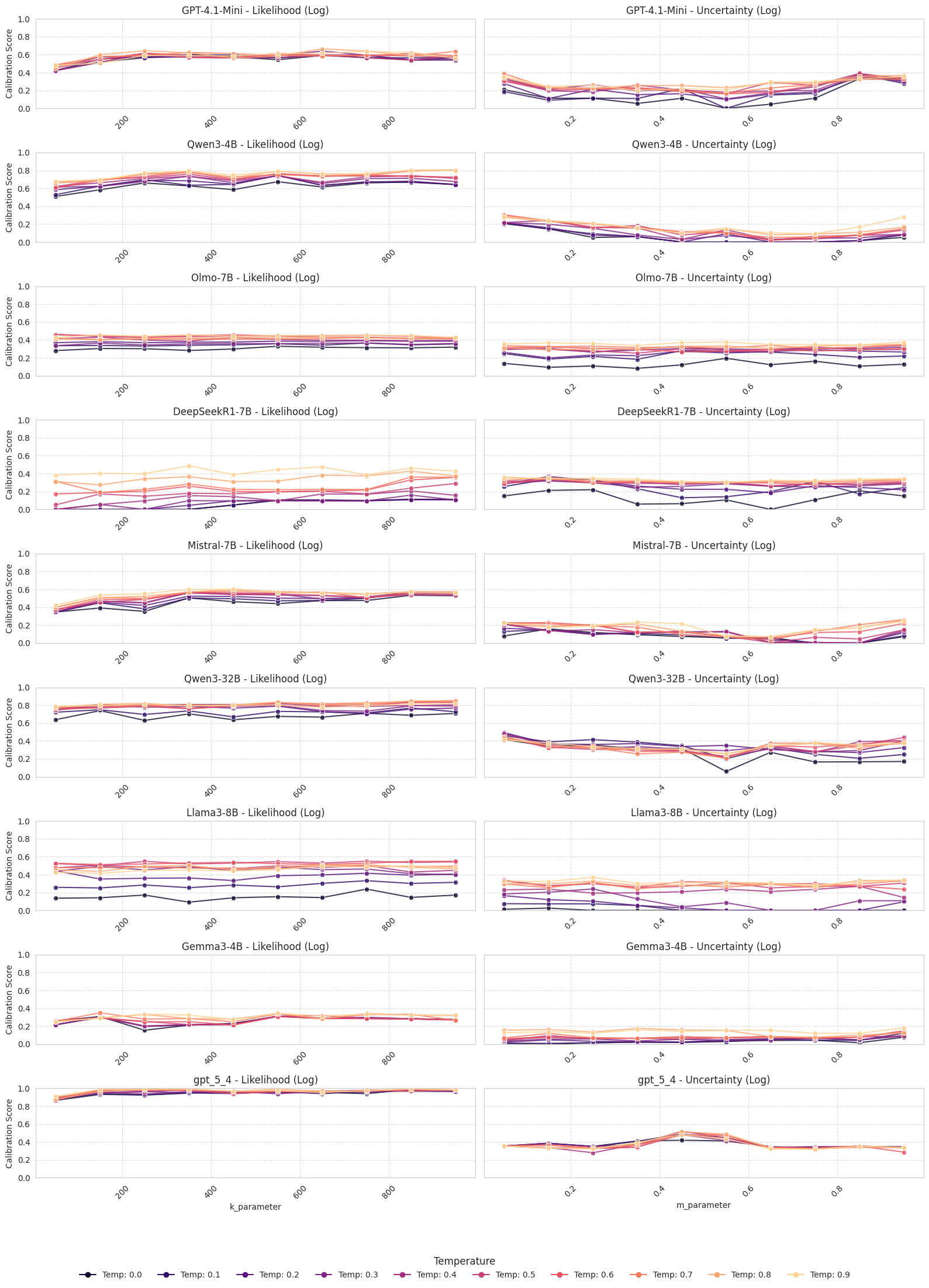}
  \end{center}
  \label{fig:dissag_logs_per_temp}
\end{figure}

\newpage
\subsection{Disaggregated consistency scores (likelihood and uncertainty) per LLM, across experiment contexts and temperatures}
\label{sec:dissag_con}

\begin{figure}[H]
  \begin{center}
  \includegraphics[width=\linewidth]{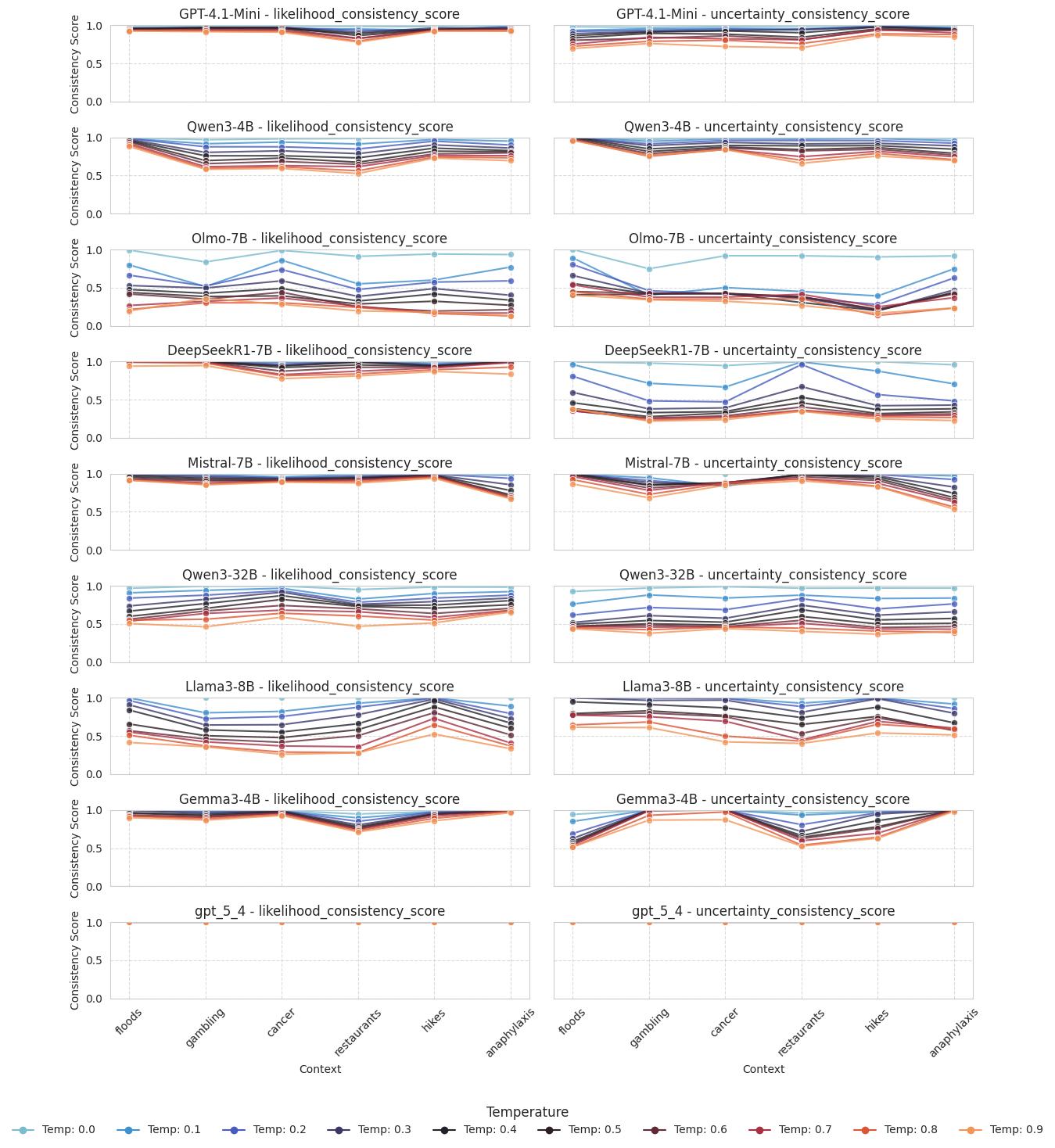}
  \end{center}
\end{figure}

\subsection{Vocabulary mappings per LLM}
\label{sec:vocab_map}

\begin{figure}[H]
  \begin{center}
  \includegraphics[width=0.85\linewidth]{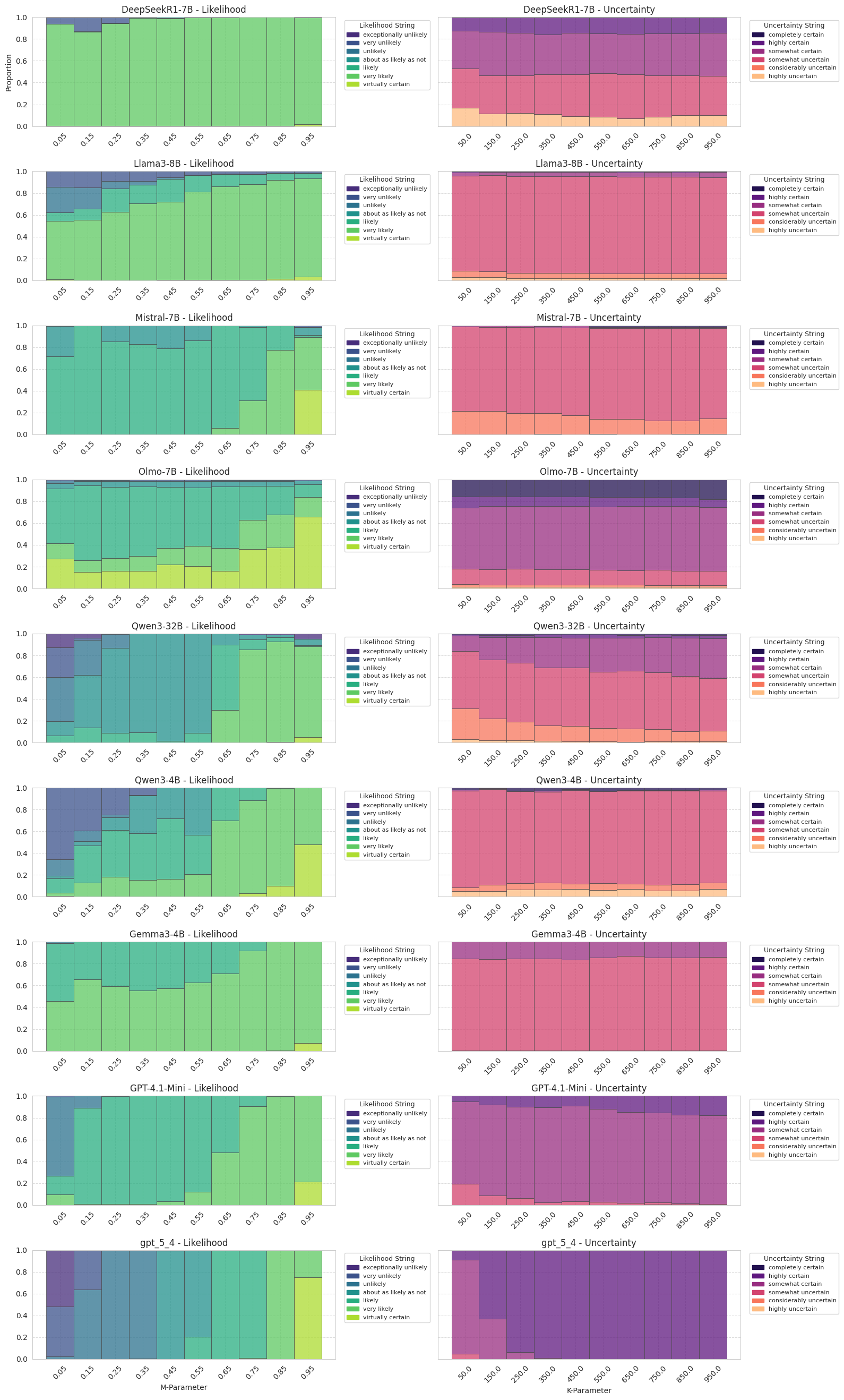}
  \end{center}
\end{figure}

\newpage
\subsection{Disaggregated vocabulary mappings per LLM, across domain contexts}
\label{sec:dissag_vocab_map}

\begin{figure}[H]
  \begin{center}
  \includegraphics[width=\linewidth]{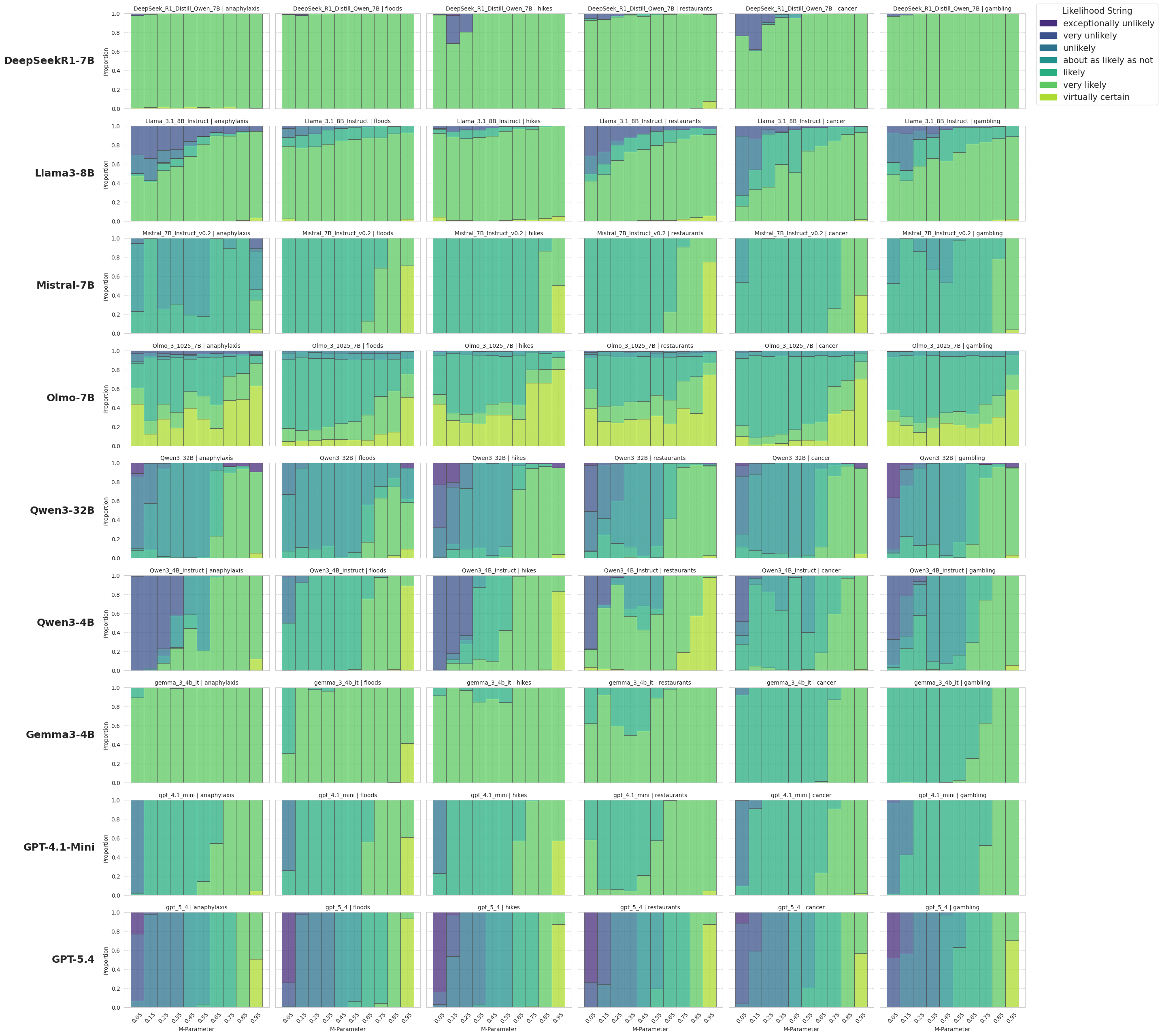}
  \end{center}
  \label{fig:dissag_vocab_map}
\end{figure}

\newpage
\bibliographystyle{plainnat}
\bibliography{bib}

\end{document}